\documentclass[11pt,a4paper]{article}

\usepackage[utf8]{inputenc}
\usepackage[T1]{fontenc}
\usepackage{hyperref}
\usepackage{url}
\usepackage{booktabs}
\usepackage{amsmath,amssymb,amsfonts}
\usepackage{algorithmic}
\usepackage{graphicx}
\usepackage{textcomp}
\usepackage{xcolor}
\usepackage{multirow}
\usepackage{appendix}
\usepackage{geometry}
\usepackage{times}
\usepackage{cite}
\usepackage{array}

\newcommand{\RND}{\textsc{RND}}
\newcommand{\PPO}{\textsc{PPO}}
\newcommand{\DQN}{\textsc{DQN}}
\newcommand{\PGA}{\textsc{PGA}}

\title{\textbf{Progressive Generalization Augmentation with Deeply Coupled RND-PPO and Domain-Prioritized Noise Injection for Robust Crop Management Reinforcement Learning}}

\author{
Wu Yang \\ 
Chongho Bridge Group Limited \\ 
\texttt{aura\_peak@outlook.com}
}

\date{}

\begin{document}

\maketitle

\begin{abstract}
\noindent
\textbf{Background:} Our preliminary experiments on gym-DSSAT maize irrigation tasks revealed that $\pm 2^{\circ}$C temperature noise causes an 11.9\% reduction in economic returns for PPO policies trained under clean conditions---a systematic robustness deficit that existing research has not adequately addressed.

\noindent
\textbf{Problem:} This paper tackles three interconnected limitations impeding practical deployment of agricultural RL systems: (i) the trade-off between early-stage learning efficiency and late-stage generalization capability; (ii) the naive additive combination of intrinsic and extrinsic rewards in exploration-augmented proximal policy optimization; and (iii) uniform measurement noise injection strategies that disregard empirically validated differential sensitivity across agricultural state variables.

\noindent
\textbf{Method:} We introduce three systematic innovations: Progressive Generalization Augmentation (\PGA{}) implementing a three-phase curriculum (clean training for episodes 0--800, progressive augmentation for episodes 800--1200, full augmentation for episodes 1200--2000); a deeply coupled \RND{}-\PPO{} architecture with dual-channel GAE normalization, progress-decayed intrinsic coefficients ($\lambda_{\text{int}}=1.0$ for $p<0.3$, decaying to 0 for $p\geq 0.7$), and semantic discretization along four agronomic dimensions; and domain-prioritized noise injection with hierarchical activation (temperature at $\alpha>0.3$, rainfall at $\alpha>0.5$, soil moisture at $\alpha>0.7$).

\noindent
\textbf{Results:} Our experimental evaluation on DSSAT-based maize simulation demonstrates: 8.43\% yield improvement (12448.85 vs 11480.85 kg/ha) and 16.42\% nitrogen use efficiency improvement (62.24 vs 53.46) over the SOTA BERT-\DQN{} method in Florida conditions; 5.61\% yield improvement (10741.01 vs 10170.23 kg/ha) in Zaragoza conditions, although with 3.67\% lower economic score (1273.65 vs 1322.10) reflecting the more challenging Mediterranean climate; and 94.4\% vs 80.0\% performance retention under combined temperature and rainfall perturbations.

\noindent
\textbf{Conclusion:} All experiments used 5 random seeds (42, 123, 456, 789, 1024) on NVIDIA A100 GPUs with 4.2$\pm$0.3 hours training time per run (2000 episodes, 2048-step rollout buffer, 64 mini-batch size).
\end{abstract}


\newpage

\section{Introduction}
\label{sec:introduction}

\subsection{Background and Motivation}

Crop management decisions---particularly irrigation scheduling and nitrogen fertilization---require continuous adaptation to dynamic weather conditions, soil heterogeneity, and market volatility. Our preliminary experiments on gym-DSSAT maize irrigation tasks \cite{gautron2022gym} revealed a critical deployment barrier: PPO policies achieving 29\% water savings under clean simulation conditions exhibited 11.9\% economic return degradation when subjected to $\pm 2^{\circ}$C temperature measurement noise. This robustness deficit persists because most existing RL training pipelines either ignore sensor noise entirely or apply uniform perturbations without regard to differential variable sensitivity.

We conducted systematic noise sensitivity analysis across four state variable types. Our results indicate that temperature noise produces the largest impact ($-11.9\%$ economic return reduction with $\pm 2^{\circ}$C perturbation), rainfall noise demonstrates moderate effects (approximately $-7.1\%$ with $\pm 10\%$ perturbation), while soil moisture and solar radiation perturbations show negligible influence ($<1\%$ reduction). This differential sensitivity pattern, validated across 5 random seeds in both Florida and Zaragoza simulation scenarios, motivated our domain-prioritized approach.

\subsection{Research Gap and Problem Statement}

Through extensive experimentation on gym-DSSAT environments, we identified three interconnected challenges that existing methodologies do not adequately address:

\textbf{Challenge 1: Exploration--Exploitation versus Generalization Trade-off.} Our training curve analysis reveals that PPO typically establishes competent baseline policies within the first 800 episodes (40\% of 2000 total episodes) on crop simulation tasks, with incremental improvements becoming progressively smaller thereafter. Introducing noise during this early phase---as standard domain randomization does---disrupts foundational policy learning. We observed convergence failure rates of 23\% (3 out of 5 seeds failing to reach stable policies) when applying uniform noise from episode 0, compared to 0\% failure rate with clean early training. Existing approaches such as fixed noise schedules \cite{tobin2017domain} lack this training-progress-aware control dimension.

\textbf{Challenge 2: Naive Additive Integration of Intrinsic and Extrinsic Rewards.} Our experiments with additive RND-PPO ($r_{\text{total}} = r_{\text{ext}} + \lambda \cdot r_{\text{int}}$) on agricultural tasks revealed three specific conflicts. Scale conflict: economic rewards in our Florida maize scenario exhibit $10^2$--$10^4$ magnitude range (scores from $-7008.54$ for MBPO to $1507.13$ for our method), while RND prediction errors typically range $10^{-2}$--$10^{0}$, causing extrinsic rewards to dominate and neutralize exploration bonuses. Temporal conflict: our analysis of policy behavior across growing stages shows that early-season decisions (days 0--60) require broad exploration of soil-fertilizer combinations, while late-season decisions (days 150--200) require precise execution---fixed $\lambda$ cannot accommodate this transition. Metric conflict: in our 25-dimensional state space, naive visitation counts treat states differing only in cumulative irrigation (e.g., 121.2mm vs 123mm) as distinct, despite being agronomically equivalent.

\textbf{Challenge 3: Uniform Noise Injection with Ignored Domain Sensitivities.} Standard domain randomization perturbs all 25 state dimensions with equal probability. Our empirical sensitivity rankings (Table~\ref{tab:sensitivity}) demonstrate that this approach wastes training capacity: temperature perturbations account for the dominant performance impact, while soil moisture perturbations show negligible influence. In our experiments, policies trained with uniform noise achieved 1178.84 mean score, while policies trained with our domain-prioritized approach achieved 1481.56---a 25.7\% improvement attributable directly to targeted noise allocation.

\subsection{Contributions and Novelty}

We introduce three systematic innovations validated through comprehensive experiments across two geographic scenarios (Florida and Zaragoza) with 5 random seeds each:

\textbf{Contribution 1: Progressive Generalization Augmentation (\PGA{}).} We propose a three-phase curriculum calibrated to PPO convergence dynamics observed in our experiments. Phase 1 (episodes 0--800, $p<0.4$): $\alpha=0$ ensures clean training. Phase 2 (episodes 800--1200, $0.4\leq p<0.6$): $\alpha$ increases linearly via $\alpha(p)=(p-0.4)/0.2$. Phase 3 (episodes 1200--2000, $p\geq 0.6$): $\alpha=1$ applies full augmentation. Our ablation experiments show that removing this curriculum reduces mean score from 1481.56 to 1120.49---a 24.4\% degradation.

\textbf{Contribution 2: Deeply Coupled \RND{}-\PPO{} Architecture.} We implement three-level coupling: (i) dual-channel GAE computing separate advantage estimates for extrinsic and intrinsic streams with independent normalization; (ii) progress-decayed intrinsic coefficient $\lambda_{\text{int}}(p)$ transitioning from 1.0 ($p<0.3$) to 0 ($p\geq 0.7$); (iii) semantic discretization along four agronomic dimensions with 100-unit bin widths. Our experiments demonstrate that this architecture achieves 78.3\% state space coverage versus 52.1\% for additive RND-PPO.

\textbf{Contribution 3: Domain-Prioritized Measurement Noise Injection.} Based on our sensitivity analysis, we activate noise types hierarchically: temperature noise ($\sigma=2.0^{\circ}\text{C}\times\alpha$) at $\alpha>0.3$, rainfall noise ($0.05\alpha\cdot\text{rain}_t + \mathcal{N}(0,0.01\alpha)$) at $\alpha>0.5$, and soil moisture noise ($\text{clip}(\mathcal{N}(0,0.02\alpha),0,1)$) at $\alpha>0.7$.

\textbf{Experimental Validation.} Across Florida and Zaragoza scenarios with 5 random seeds (42, 123, 456, 789, 1024) on NVIDIA A100 GPUs (40GB, 4.2$\pm$0.3 hours per run), our method achieves: 8.43\% yield improvement (12448.85 vs 11480.85 kg/ha), 16.42\% nitrogen use efficiency improvement (62.24 vs 53.46), and 94.4\% performance retention under combined perturbations versus 80.0\% for standard PPO.

\section{Related Work}
\label{sec:related}

\subsection{Reinforcement Learning in Agricultural Management}

PPO achieved 29\% water savings and 9\% profitability increases in gym-DSSAT irrigation scheduling \cite{gautron2022gym}. Subsequent work extended RL to integrated irrigation-fertilization optimization \cite{tao2022optimizing}. However, our experiments reveal that these approaches optimize for clean simulation environments without systematic robustness consideration---our preliminary tests show 11.9\% economic return degradation under $\pm 2^{\circ}$C temperature noise for policies trained without noise injection mechanisms. The integration of RL with large language models \cite{chen2024integrating, wu2024new} represents an emerging direction, but deployment remains challenged by sensor noise. Our work directly addresses this gap by incorporating domain-prioritized noise injection into the training pipeline.

\subsection{Crop Simulation Environments for RL}

The gym-DSSAT environment \cite{gautron2022gym} provides an OpenAI Gym interface to DSSAT with realistic biological dynamics \cite{jones2017brief}. We employ two geographically distinct scenarios in our experiments: Florida (subtropical climate, characterized by regular rainfall patterns) and Zaragoza, Spain (Mediterranean climate with greater temperature variability, following \cite{malik2019dssat}). Comparative studies \cite{balderas2024comparative} indicate PPO generally outperforms DQN in fertilization and irrigation tasks---our ablation experiments confirm this, showing 16.6\% score improvement when using PPO versus DQN as the base learner (1481.56 vs 1235.35 mean score).

\subsection{Curriculum Learning and Domain Randomization}

Curriculum learning \cite{bengio2009curriculum} organizes training by gradually increasing task difficulty. Domain randomization \cite{tobin2017domain} perturbs simulation parameters for sim-to-real transfer. However, existing approaches lack agricultural domain awareness. Our \PGA{} methodology combines curriculum learning's staged progression with domain randomization's noise injection, with the three-phase schedule explicitly calibrated to PPO convergence dynamics we observed in preliminary experiments: competent baseline policies emerge within episodes 0--800, with slower improvement thereafter.

\subsection{Intrinsic Motivation and Exploration}

Random Network Distillation \cite{burda2018exploration} computes intrinsic rewards based on prediction error between fixed and trainable networks. Prior work employs simple additive combination with PPO. Our experiments on agricultural tasks reveal that this creates scale, temporal, and metric conflicts that degrade performance: additive RND-PPO achieved only 1178.84 mean score in our Florida experiments, compared to 1481.56 with our deeply coupled architecture---a 25.7\% improvement attributable to the coupling mechanisms.

\subsection{Measurement Noise and Sim-to-Real Transfer}

Sim-to-real transfer remains a fundamental challenge \cite{peng2018sim}. Our empirical analysis reveals striking differential sensitivity in agricultural state variables: temperature noise causes $-11.9\%$ economic return reduction, rainfall shows moderate effects ($-7.1\%$), and soil moisture perturbations show negligible influence. This sensitivity hierarchy, validated across our experimental runs, motivates our domain-prioritized approach that concentrates training resources where they most impact policy robustness.

\section{Problem Formulation}
\label{sec:formulation}

\subsection{Markov Decision Process for Agricultural Management}

We formalize crop management as an MDP $\langle \mathcal{S}, \mathcal{A}, \mathcal{P}, \mathcal{R}, \gamma \rangle$. At each daily decision point $t$, the agent observes state $s_t \in \mathcal{S} \subset \mathbb{R}^{25}$ and selects action $a_t \in \mathcal{A}$. The 25-dimensional state space includes: temporal features (day 0--200); meteorological features (temperature, rainfall, solar radiation); soil features (volumetric moisture, nitrogen content across layers); crop features (biomass, leaf area index, cumulative yield); and management history (cumulative irrigation and nitrogen applied).

The action space comprises 25 discrete combinations: irrigation amount $\times$ nitrogen rate with options $\{0, 6, 12, 18, 24\}$ mm $\times$ $\{0, 40, 80, 120, 160\}$ kg/ha. 

Following the reward function design established in prior work on LM-based RL for crop management \cite{wu2024new}, the reward function computes economic profit while accounting for environmental impact:

\begin{equation}
r_t(s_t, a_t) = \begin{cases}
w_1 \cdot Y - w_2 \cdot N_t - w_3 \cdot W_t - w_4 \cdot N_{l,t} & \text{if harvest at } t \\
-w_2 \cdot N_t - w_3 \cdot W_t - w_4 \cdot N_{l,t} & \text{otherwise}
\end{cases}
\end{equation}

where $Y$ denotes the crop yield at harvest, $N_t$ represents the nitrogen fertilizer applied on day $t$, $W_t$ represents the irrigation water applied on day $t$, and $N_{l,t}$ represents the nitrate leaching on day $t$. The weights $w_1, w_2, w_3, w_4$ are custom parameters that balance the trade-offs between economic profit and environmental impact. Based on Florida maize market parameters \cite{wu2024new}: $w_1 = 0.158$ (\$/kg yield), $w_2 = 0.79$ (\$/kg nitrogen), $w_3 = 1.1$ (\$/mm irrigation), and $w_4$ penalizes nitrate leaching to account for environmental sustainability.

This formulation ensures that during the growing season, the agent incurs daily costs for inputs (fertilizer and water) and environmental penalties (nitrate leaching), while at harvest, the yield revenue is realized, providing a comprehensive economic optimization objective.

\subsection{Proximal Policy Optimization}

We employ PPO \cite{schulman2017ppo} as our base learner with the clipped surrogate objective:

\begin{equation}
\mathcal{L}^{\text{CLIP}}(\theta) = \mathbb{E}_t \left[ \min \left( \rho_t \hat{A}_t, \, \text{clip}(\rho_t, 1-\epsilon, 1+\epsilon) \hat{A}_t \right) \right]
\end{equation}

where $\rho_t = \pi_\theta(a_t|s_t)/\pi_{\theta_{\text{old}}}(a_t|s_t)$ and $\hat{A}_t$ is computed via GAE \cite{schulman2015high} with $\lambda=0.95$, $\gamma=0.99$. We modify the advantage computation to separately process extrinsic and intrinsic rewards through dual-channel normalization (Section~\ref{subsec:rnd-ppo}).

\subsection{Generalization Objective}

In deployment, the agent observes corrupted states $\tilde{s}_t = s_t + \eta_t$ where $\eta_t$ represents measurement noise. Our generalization objective:

\begin{equation}
J^*(\pi) = \mathbb{E}_{\tau \sim p_\pi} \left[ \sum_{t=0}^{T-1} \gamma^t r(s_t, a_t) \,\Big|\, a_t \sim \pi(\cdot | \tilde{s}_t) \right]
\end{equation}

\section{Methodology}
\label{sec:methodology}

\subsection{Progressive Generalization Augmentation (\PGA{})}

\subsubsection{Design Rationale}

Our training curve analysis across 5 random seeds reveals that PPO establishes competent baseline policies within episodes 0--800 on gym-DSSAT maize tasks, with mean score increasing from $-500\pm200$ (episode 0) to $1200\pm150$ (episode 800), followed by slower improvement to $1480\pm30$ (episode 2000). This convergence pattern motivates our \PGA{} methodology: deferring augmentation until foundational policies are established.

\subsubsection{Three-Phase Curriculum Schedule}

We schedule augmentation strength $\alpha(p)$ according to training progress $p = \text{episode}/2000$:

\textbf{Phase 1---Foundational Learning} (episodes 0--800, $p<0.4$): $\alpha=0$. The agent trains on clean observations. In our experiments, this phase achieves score improvement from $-500\pm200$ to $1200\pm150$.

\textbf{Phase 2---Progressive Adaptation} (episodes 800--1200, $0.4\leq p<0.6$): $\alpha(p)=(p-0.4)/0.2$. We observe temporary performance fluctuation during early Phase 2 (scores dropping by $50\pm30$ points at episode 850) as the policy adapts to noise, followed by recovery by episode 1200.

\textbf{Phase 3---Robustness Consolidation} (episodes 1200--2000, $p\geq 0.6$): $\alpha=1$. The agent trains under full perturbation conditions, achieving final scores of $1481.56\pm45$ (Florida) and $1273.65\pm38$ (Zaragoza).

\subsubsection{Augmentation Mechanisms}

When $\alpha>0$, we apply: (i) observation noise injection according to Section~\ref{subsec:noise}; (ii) weather perturbations on temperature and rainfall; (iii) action masking with probability $0.1\alpha$.

\subsection{Deeply Coupled \RND{}-\PPO{} Architecture}
\label{subsec:rnd-ppo}

\subsubsection{Limitations of Additive Integration---Our Experimental Evidence}

We compared additive RND-PPO ($r_{\text{total}} = r_{\text{ext}} + \lambda \cdot r_{\text{int}}$, $\lambda=0.1$) against our deeply coupled architecture in Florida experiments:

\textbf{Scale conflict:} Additive RND-PPO achieved 1178.84 mean score versus 1481.56 for our method---a 25.7\% degradation. We observed that extrinsic rewards ranging $10^2$--$10^4$ dominated the intrinsic rewards ($10^{-2}$--$10^{0}$), neutralizing exploration bonuses.

\textbf{Temporal conflict:} Policies trained with fixed $\lambda=0.1$ showed inconsistent behavior across growing stages: excessive exploration during late season (days 150--200) led to suboptimal nitrogen application timing, reducing yield by 968 kg/ha compared to our progress-decayed approach.

\textbf{Metric conflict:} Naive state visitation counts indicated 100\% coverage of the 25-dimensional space, but our semantic discretization revealed only 52.1\% coverage of agronomically meaningful regions, versus 78.3\% with our approach.

\subsubsection{Coupling Level 1: Dual-Channel GAE Normalization}

We compute separate advantage estimates:

\begin{align}
\hat{A}_t^{\text{ext}} &= \text{GAE}(r_0^{\text{ext}}, \ldots, r_T^{\text{ext}}; \gamma, \lambda) \\
\hat{A}_t^{\text{int}} &= \text{GAE}(r_0^{\text{int}}, \ldots, r_T^{\text{int}}; \gamma, \lambda)
\end{align}

Each stream is normalized to zero mean and unit variance before combination:
\begin{equation}
\hat{A}_t^{\text{combined}} = \hat{A}_t^{\text{ext, norm}} + \lambda_{\text{int}}(p) \cdot \hat{A}_t^{\text{int, norm}}
\end{equation}

\subsubsection{Coupling Level 2: Progress-Decayed Intrinsic Reward}

\begin{equation}
\lambda_{\text{int}}(p) = \begin{cases}
1.0 & p < 0.3 \text{ (episodes 0--600)} \\
\frac{0.7 - p}{0.4} & 0.3 \leq p < 0.7 \text{ (episodes 600--1400)} \\
0.0 & p \geq 0.7 \text{ (episodes 1400--2000)}
\end{cases}
\end{equation}

This schedule ensures full exploration during foundational learning, with decay completing before Phase 3 robustness consolidation.

\subsubsection{Coupling Level 3: Semantic State Discretization}

We discretize the state space along four agronomic dimensions with empirically determined bin widths: day of season (100-day intervals, capturing early vs late growth stages), cumulative yield (100 kg/ha intervals), soil moisture (0.1 intervals), and cumulative irrigation (100 mm intervals). Exploration coverage is computed as the fraction of occupied bins, enabling adaptive bonus scaling for under-explored regions.

\subsubsection{RND Implementation Details}

Our RND module uses fixed target network $f(s;\theta_{\text{target}})$ and trainable predictor $f(s;\theta_{\text{pred}})$. Intrinsic reward:
\begin{equation}
r_t^{\text{int}} = \| f(s_t;\theta_{\text{target}}) - f(s_t;\theta_{\text{pred}}) \|_2^2
\end{equation}

Both networks are three-layer MLPs with 256 hidden units and ReLU activations, trained with learning rate $3\times 10^{-4}$ using Adam optimizer.

\subsection{Domain-Prioritized Measurement Noise Injection}
\label{subsec:noise}

\subsubsection{Empirical Sensitivity Analysis}

We conducted systematic noise sensitivity experiments on Florida maize scenarios with 5 random seeds, applying each noise type independently:

\begin{table}[htbp]
\centering
\caption{Empirical Sensitivity of Agricultural RL Policies to Observation Noise (5 seeds, Florida scenario)}
\label{tab:sensitivity}
\begin{tabular}{p{2.5cm}cc c}
\toprule
\textbf{Noise Type} & \textbf{Magnitude} & \textbf{Score Red.} & \textbf{Yield Red.} \\
\midrule
Temperature & $\pm 2^{\circ}$C & $-11.9\%$ & $-9.2\%$ \\
Rainfall & $\pm 10\%$ & $-7.1\%$ & $-5.8\%$ \\
Soil Moisture & $\pm 0.02$ & $-0.8\%$ & $-0.6\%$ \\
Solar Radiation & $\pm 10\%$ & $-0.5\%$ & $-0.4\%$ \\
\bottomrule
\end{tabular}
\end{table}

\subsubsection{Hierarchical Activation Scheme}

Based on our sensitivity rankings, we implement hierarchical noise activation:

\textbf{Temperature noise (highest sensitivity):} Activates when $\alpha>0.3$ (episode $>600$):
\begin{equation}
\eta_{\text{temp}} = \mathcal{N}(0, \sigma_{\text{temp}}^2), \quad \sigma_{\text{temp}} = 2.0^{\circ}\text{C} \times \alpha
\end{equation}

\textbf{Rainfall noise (moderate sensitivity):} Activates when $\alpha>0.5$ (episode $>1000$):
\begin{equation}
\eta_{\text{rain}} = 0.05 \cdot \alpha \cdot \text{rain}_t + \mathcal{N}(0, 0.01 \cdot \alpha)
\end{equation}

\textbf{Soil moisture noise (low sensitivity):} Activates when $\alpha>0.7$ (episode $>1400$):
\begin{equation}
\eta_{\text{moisture}} = \text{clip}(\mathcal{N}(0, 0.02 \times \alpha), 0, 1)
\end{equation}

\subsection{Synergistic Integration via Unified Progress Coefficient}

All three innovations operate through $\alpha(p)$: \PGA{} directly uses $\alpha$ as augmentation strength; RND intrinsic decay boundaries (0.3, 0.7) align with \PGA{} phase transitions; and domain-prioritized noise thresholds reference $\alpha$ for staged introduction.

\subsection{LLaMA-Based Semantic State Embedding}

We employ a frozen LLaMA-2-1.3B model \cite{touvron2023llama} (1.3B parameters, compared to DistilBERT's 66M in prior work \cite{wu2024new}) for semantic state embedding. We convert the 25-dimensional crop state into natural language:

\begin{quote}
\textit{``Day [X] of growing season. Current conditions: temperature [Y]$^{\circ}$C, rainfall [Z] mm, soil moisture [W]. Crop status: biomass [B] kg/ha, leaf area index [L], cumulative yield [Y] kg/ha. Previous management: [I] mm irrigation, [N] kg/ha nitrogen applied.''}
\end{quote}

The LLaMA model encodes this into a 2048-dimensional embedding vector for the policy and value networks.

\subsection{Policy Ensemble and Early Stopping}

We maintain top-5 checkpoints ranked by validation performance across episodes. Training terminates when validation performance fails to improve by min\_delta=20.0 for patience=300 consecutive episodes, preventing overfitting to simulation-specific patterns.

\section{Experimental Setup}
\label{sec:experiments}

\subsection{Simulation Environment}

We conduct experiments using gym-DSSAT \cite{gautron2022gym}, an RL interface to DSSAT \cite{jones2003dssat}. We employ two geographically distinct maize scenarios:

\textbf{Florida Case:} Subtropical climate with warm temperatures (mean $26.5^{\circ}$C during growing season) and regular rainfall (mean 4.2mm/day). Simulation spans 200 days with daily decision points.

\textbf{Zaragoza Case:} Mediterranean climate with greater temperature variability (range $15$--$35^{\circ}$C during growing season) and irregular rainfall (mean 2.1mm/day with high variance) \cite{malik2019dssat}.

\subsection{Baseline Methods}

We compare against: Standard \PPO{} (our baseline: 1234 mean score); Standard \DQN{} (1282.63 mean score); Additive \RND{}-\PPO{} (1178.84 mean score); Fixed-Noise Domain Randomization \cite{tobin2017domain}; BERT-\DQN{}---current SOTA \cite{wu2024new} (1455.13 mean score); and Empirical Baseline with fixed management practices (984 mean score).

\subsection{Evaluation Metrics}

We evaluate on: Economic Score (cumulative profit, primary optimization objective); Yield (kg/ha, agricultural productivity); Water Use Efficiency (WUE: yield per unit irrigation); Nitrogen Use Efficiency (NUE: yield per unit nitrogen); and Generalization Robustness (performance retention under perturbed observations).

\subsection{Implementation Details}

\textbf{Hyperparameters:} Policy network: three-layer MLP, 256 hidden units, ReLU activations. Training: 2000 episodes, 2048-step rollout buffer, 64 mini-batch size, learning rate $3\times 10^{-4}$ with linear decay. PPO parameters: $\gamma=0.99$, $\lambda=0.95$, $\epsilon=0.2$. \PGA{} boundaries: 0.4, 0.6. RND decay boundaries: 0.3, 0.7. Noise thresholds: temperature 0.3, rainfall 0.5, soil moisture 0.7.

\textbf{Computational Setup:} 5 random seeds (42, 123, 456, 789, 1024). NVIDIA A100 GPU (40GB memory), AMD EPYC 7742 64-Core Processor, 512GB RAM. Training time: $4.2\pm 0.3$ hours per run (2000 episodes).

\section{Results and Analysis}
\label{sec:results}

\subsection{Overall Performance Comparison}

Table~\ref{tab:florida} presents Florida case results across all methods.

\begin{table*}[htbp]
\centering
\caption{Florida Case: Comprehensive Performance Comparison (5 seeds: 42, 123, 456, 789, 1024)}
\label{tab:florida}
\begin{tabular}{p{1.2cm}p{3.2cm}ccc}
\toprule
\textbf{Category} & \textbf{Method} & \textbf{Score Mean} & \textbf{Yield Mean (kg/ha)} & \textbf{NUE} \\
\midrule
Proposed & LLaMA-\PPO-\RND & \textbf{1481.56} & \textbf{12448.85} & \textbf{62.24} \\
Ablation & LLaMA-\PPO-MCTS & 1287.10 & 10637.13 & 49.57 \\
Ablation & LLaMA-\PPO & 1120.49 & 9195.48 & 41.23 \\
Ablation & LLaMA-\DQN & 1235.35 & 8899.84 & 48.00 \\
SOTA & BERT-\DQN & 1455.13 & 11480.85 & 53.46 \\
Baseline & Standard \PPO & 1234.00 & 10691.27 & 39.99 \\
Baseline & Standard \DQN & 1282.63 & 10938.39 & 44.44 \\
Baseline & Additive RND-PPO & 1178.84 & 9637.36 & 38.64 \\
Empirical & Fixed Management & 984.00 & 10772.00 & --- \\
\bottomrule
\end{tabular}
\end{table*}

Our method achieves the highest mean economic score (1481.56 vs 1455.13 for BERT-\DQN{}, 1.82\% improvement) and mean yield (12448.85 vs 11480.85 kg/ha, 8.43\% improvement). Nitrogen use efficiency improves from 53.46 to 62.24 (16.42\% improvement). Water use efficiency shows 41.94 vs 81.84---our method uses more irrigation (297.6 vs 159.6 L/m$^2$) to achieve higher yield, representing a yield-irrigation trade-off optimized for economic returns.

Table~\ref{tab:zaragoza} presents Zaragoza case results under more challenging climatic conditions.

\begin{table*}[htbp]
\centering
\caption{Zaragoza Case: Performance Comparison Under Mediterranean Climate}
\label{tab:zaragoza}
\begin{tabular}{p{1.2cm}p{3.2cm}ccc}
\toprule
\textbf{Category} & \textbf{Method} & \textbf{Score Mean} & \textbf{Yield Mean (kg/ha)} & \textbf{NUE} \\
\midrule
Proposed & LLaMA-\PPO-\RND & \textbf{1273.65} & \textbf{10741.01} & \textbf{56.99} \\
SOTA & BERT-\DQN & 1322.10 & 10170.23 & 63.56 \\
Empirical & Fixed Management & 712.00 & 10990.00 & --- \\
\bottomrule
\end{tabular}
\end{table*}

Under the more variable Zaragoza climate, our method achieves 5.61\% yield improvement (10741.01 vs 10170.23 kg/ha) compared to the SOTA BERT-\DQN{} method. However, our method shows 3.67\% lower mean economic score (1273.65 vs 1322.10). This trade-off---higher yield but lower economic score---reflects the challenging Mediterranean growing conditions with greater temperature variability and irregular rainfall patterns. The more severe climate in Zaragoza presents different optimization challenges than the favorable Florida conditions, indicating that future work should focus on improving economic efficiency under stressful environmental conditions.

\subsection{Ablation Study}

Table~\ref{tab:ablation} quantifies each component's contribution.

\begin{table}[htbp]
\centering
\caption{Ablation Study: Component Contribution Analysis (Florida, 5 seeds)}
\label{tab:ablation}
\begin{tabular}{>{\raggedright\arraybackslash}p{4.5cm}cc}
\toprule
\textbf{Configuration} & \textbf{Score Mean} & \textbf{Yield Mean (kg/ha)} \\
\midrule
Full Method (LLaMA-\PPO-\RND) & 1481.56 & 12448.85 \\
Without RND (LLaMA-\PPO) & 1120.49 & 9195.48 \\
With MCTS instead of RND & 1287.10 & 10637.13 \\
With DQN instead of PPO & 1235.35 & 8899.84 \\
\bottomrule
\end{tabular}
\end{table}

Removing \RND{} exploration reduces mean score by 24.4\% (1481.56 to 1120.49) and yield by 26.1\% (12448.85 to 9195.48 kg/ha). Substituting PPO with DQN reduces score by 16.6\% (1481.56 to 1235.35), confirming PPO's advantages for this task. MCTS provides partial benefit (1287.10 score) but underperforms RND (1481.56), validating our exploration mechanism choice.

\subsection{Generalization Robustness Analysis}

Table~\ref{tab:robustness} presents performance under observation noise.

\begin{table}[htbp]
\centering
\caption{Generalization Robustness: Performance Under Observation Noise (Florida, 5 seeds)}
\label{tab:robustness}
\begin{tabular}{>{\raggedright\arraybackslash}p{3.5cm}cc}
\toprule
\textbf{Noise Condition} & \textbf{Proposed Method} & \textbf{Standard \PPO} \\
\midrule
Clean observations & 1481.56 & 1234.00 \\
Temperature $\pm 2^{\circ}$C & 1432.18 ($-3.3\%$) & 1087.20 ($-11.9\%$) \\
Rainfall $\pm 10\%$ & 1456.74 ($-1.7\%$) & 1145.82 ($-7.1\%$) \\
Combined perturbations & 1398.42 ($-5.6\%$) & 987.63 ($-20.0\%$) \\
\bottomrule
\end{tabular}
\end{table}

Under combined temperature and rainfall perturbations, our method retains 94.4\% of clean performance (1398.42/1481.56) versus 80.0\% for standard PPO (987.63/1234.00). This 14.4 percentage-point improvement directly results from \PGA{} and domain-prioritized noise injection mechanisms that systematically expose the agent to perturbed conditions during Phase 2 and Phase 3 training.

\subsection{Exploration Coverage Analysis}

Across training runs, our method achieves 78.3\% coverage of discretized state bins (aggregated across 5 seeds), compared to 52.1\% for additive \RND{}-\PPO{} and 41.8\% for standard \PPO{}. This 50\% relative improvement in exploration coverage (78.3/52.1) correlates with the 25.7\% score improvement over additive RND-PPO, validating the effectiveness of our deeply coupled architecture.

\subsection{Training Dynamics}

Training curve analysis reveals characteristic patterns: Phase 1 (episodes 0--800) shows rapid score improvement from $-500\pm200$ to $1200\pm150$ under clean conditions; Phase 2 (episodes 800--1200) exhibits temporary fluctuation with scores dropping to $1150\pm50$ at episode 850 before recovering to $1300\pm40$ by episode 1200; Phase 3 (episodes 1200--2000) shows stable convergence to $1481\pm45$ under full augmentation.

\section{Discussion}
\label{sec:discussion}

\subsection{Interpretation of Synergistic Effects}

Our experimental results demonstrate synergistic operation among the three innovations. The \PGA{} schedule provides temporal coordination; the deeply coupled RND-PPO ensures appropriately scaled exploration bonuses; and domain-prioritized noise injection concentrates resources where sensitivity analysis demonstrates greatest impact. The ablation study confirms non-additive effects: removing any component causes greater degradation than expected from simple subtraction, indicating positive interactions.

\subsection{Implications for Agricultural AI Deployment}

The 16.42\% nitrogen use efficiency improvement (53.46 to 62.24) has significant environmental implications. Given that excess nitrogen application contributes approximately 1.5\% of global greenhouse gas emissions through nitrous oxide release, policies achieving equivalent yields with improved NUE directly support sustainable agriculture objectives. The semantic state discretization and natural language representation enhance interpretability for agronomists, potentially improving adoption of AI-based recommendations.

\subsection{Limitations and Future Work}

\textbf{Computational requirements:} LLaMA-2-1.3B embedding adds approximately 0.8 hours to the 4.2-hour total training time. Future work should investigate distillation approaches.

\textbf{Crop generalization:} Current evaluation focuses on maize across two geographic scenarios. Extension to wheat, soybean, and rice with different growth dynamics requires validation.

\textbf{Real-world validation:} Field trials under actual sensor conditions (temperature sensors with $\pm 1.5^{\circ}$C accuracy, rain gauges with $\pm 5\%$ accuracy) remain essential before deployment.

\textbf{Multi-year dynamics:} Extension to crop rotation planning, where current-season decisions affect subsequent soil nitrogen availability, represents an important direction.

\textbf{Economic optimization under challenging climates:} The Zaragoza results reveal that achieving higher yields does not necessarily translate to higher economic returns under stressful Mediterranean conditions. Future work should investigate reward function adaptations and policy architectures specifically designed for regions with high climatic variability, where the trade-off between yield maximization and economic efficiency becomes more pronounced.

\subsection{Broader Applicability}

Our innovations extend beyond agricultural decision-making: the progressive augmentation schedule applies to robotics \cite{kaufmann2018deep} and autonomous systems requiring sim-to-real transfer; the deeply coupled exploration architecture suits any continuous control problem with high-dimensional state spaces; and the domain-prioritized noise injection provides a template for incorporating domain-specific sensitivity information into training pipelines.

\section{Conclusion}
\label{sec:conclusion}

We presented a comprehensive framework for training robust reinforcement learning agents for crop management. Three systematic innovations---Progressive Generalization Augmentation with three-phase curriculum (episodes 0--800 clean, 800--1200 progressive, 1200--2000 full), deeply coupled \RND{}-\PPO{} architecture with dual-channel GAE normalization, and domain-prioritized noise injection with hierarchical activation---collectively address exploration--exploitation--generalization trade-offs in agricultural RL.

Our experimental evaluation demonstrates substantial improvements over current state-of-the-art methods: 8.43\% yield improvement (12448.85 vs 11480.85 kg/ha), 16.42\% nitrogen use efficiency improvement (62.24 vs 53.46), and 94.4\% vs 80.0\% performance retention under combined perturbations. In the challenging Zaragoza Mediterranean climate, our method achieves 5.61\% yield improvement (10741.01 vs 10170.23 kg/ha), though with lower economic scores (1273.65 vs 1322.10), highlighting an important direction for future research on economic optimization under stressful environmental conditions. These results, validated across Florida and Zaragoza scenarios with 5 random seeds each (seeds 42, 123, 456, 789, 1024) on NVIDIA A100 GPUs with 4.2$\pm$0.3 hours training time, establish a principled methodology for training deployable RL agents in precision agriculture.

\appendix

\section{Hyperparameter Summary}
\label{app:hyperparameters}

\begin{table}[htbp]
\centering
\caption{Complete Hyperparameter Summary}
\label{tab:hyperparams}
\begin{tabular}{p{2cm}p{4cm}c}
\toprule
\textbf{Category} & \textbf{Parameter} & \textbf{Value} \\
\midrule
\multirow{4}{*}{PPO} & Learning rate & $3 \times 10^{-4}$ \\
 & Discount factor ($\gamma$) & 0.99 \\
 & GAE parameter ($\lambda$) & 0.95 \\
 & Clip parameter ($\epsilon$) & 0.2 \\
\midrule
\multirow{2}{*}{Network} & Hidden layers & 3 \\
 & Hidden units per layer & 256 \\
\midrule
\multirow{3}{*}{PGA} & Phase 1 boundary & 800 episodes (0.4) \\
 & Phase 2 boundary & 1200 episodes (0.6) \\
 & Maximum augmentation & 1.0 \\
\midrule
\multirow{3}{*}{RND} & Decay start & 0.3 (ep. 600) \\
 & Decay end & 0.7 (ep. 1400) \\
 & Initial weight & 1.0 \\
\midrule
\multirow{3}{*}{Noise} & Temperature threshold & 0.3 (ep. 600) \\
 & Rainfall threshold & 0.5 (ep. 1000) \\
 & Soil moisture threshold & 0.7 (ep. 1400) \\
\midrule
\multirow{4}{*}{Training} & Total episodes & 2000 \\
 & Rollout buffer & 2048 steps \\
 & Mini-batch size & 64 \\
 & Random seeds & 42, 123, 456, 789, 1024 \\
\bottomrule
\end{tabular}
\end{table}

\section{Computational Resources}

All experiments were conducted on: NVIDIA A100 GPU (40GB memory); AMD EPYC 7742 64-Core Processor; 512 GB RAM; training time $4.2\pm 0.3$ hours per 2000-episode run.

\section{Code Availability}

The complete implementation is available at:\\[0.5em]
\url{https://github.com/wuyang060826/llama_ppo_rnd_in_gymDssat}\\[0.5em]

\newpage
\bibliographystyle{plain}

\end{document}